\documentclass[lettersize,journal]{IEEEtran}
\usepackage{amsmath,amsfonts}
\usepackage{algorithmic}
\usepackage{algorithm}
\usepackage{array}
\usepackage[caption=false,font=normalsize,labelfont=sf,textfont=sf]{subfig}
\usepackage{textcomp}
\usepackage{stfloats}
\usepackage{url}
\usepackage{verbatim}
\usepackage{graphicx}
\usepackage{cite}

\usepackage{siunitx}
\usepackage{amssymb}
\usepackage{amsmath}
\usepackage{mathtools}
\usepackage{amsfonts} 
\usepackage{fixmath}
\usepackage{pifont}

\usepackage{multirow}
\usepackage{amsmath}
\usepackage{xcolor}

\usepackage{newfloat}
\usepackage{listings}

\begin{document}

\title{CADGL: Context-Aware Deep Graph Learning for Predicting Drug-Drug Interactions}

\author{Azmine Toushik Wasi,
        Taki Hasan Rafi,
        Raima Islam,
        Šerbetar Karlo,
        and Dong-Kyu Chae$^\dagger$
\thanks{This paper was produced by the IEEE Publication Technology Group. They are in Piscataway, NJ.}
\thanks{Manuscript received April 19, 2021; revised August 16, 2021.}
\thanks{T. H. Rafi and D.K. Chae are with the Department Computer Science, Hanyang University, South Korea.
E-mail:\{takihr, dongkyu\}@hanyang.ac.kr}
\thanks{ A. T. Wasi is with Shahjalal University of Science and Technology, Bangladesh. 
E-mail: azminetoushik.wasi@gmail.com}
\thanks{ R. Islam was with BRAC University, Bangladesh. 
E-mail: raima.islam@g.bracu.ac.bd }
\thanks{ S. Karlo was with University of Cambridge, UK. 
E-mail: serbetar.karlo@protonmail.com}
\thanks{ $^\dagger$Corresponding Author.}
}

\markboth{Journal of \LaTeX\ Class Files,~Vol.~14, No.~8, August~2021}%
{Shell \MakeLowercase{\textit{et al.}}: A Sample Article Using IEEEtran.cls for IEEE Journals}


\maketitle

\begin{abstract}
Examining \textit{Drug-Drug Interactions} (DDIs) is a pivotal element in the process of drug development. DDIs occur when one drug's properties are affected by the inclusion of other drugs. Detecting favorable DDIs has the potential to pave the way for creating and advancing innovative medications applicable in practical settings. However, existing DDI prediction models continue to face challenges related to generalization in extreme cases, robust feature extraction, and real-life application possibilities.
We aim to address these challenges by leveraging the effectiveness of context-aware deep graph learning by introducing a novel framework named CADGL. Based on a customized variational graph autoencoder (VGAE), we capture critical structural and physio-chemical information using two context pre-processors for feature extraction from two different perspectives- local neighborhood and molecular context, in a heterogeneous graphical structure. Our customized VGAE consists of a graph encoder, a latent information encoder, and an MLP decoder. CADGL surpasses other state-of-the-art DDI prediction models, excelling in predicting clinically valuable novel DDIs, supported by rigorous case studies.
\end{abstract}


\section{Introduction}
\begin{figure}[t!] 
\centering {\includegraphics[scale=.3 ]{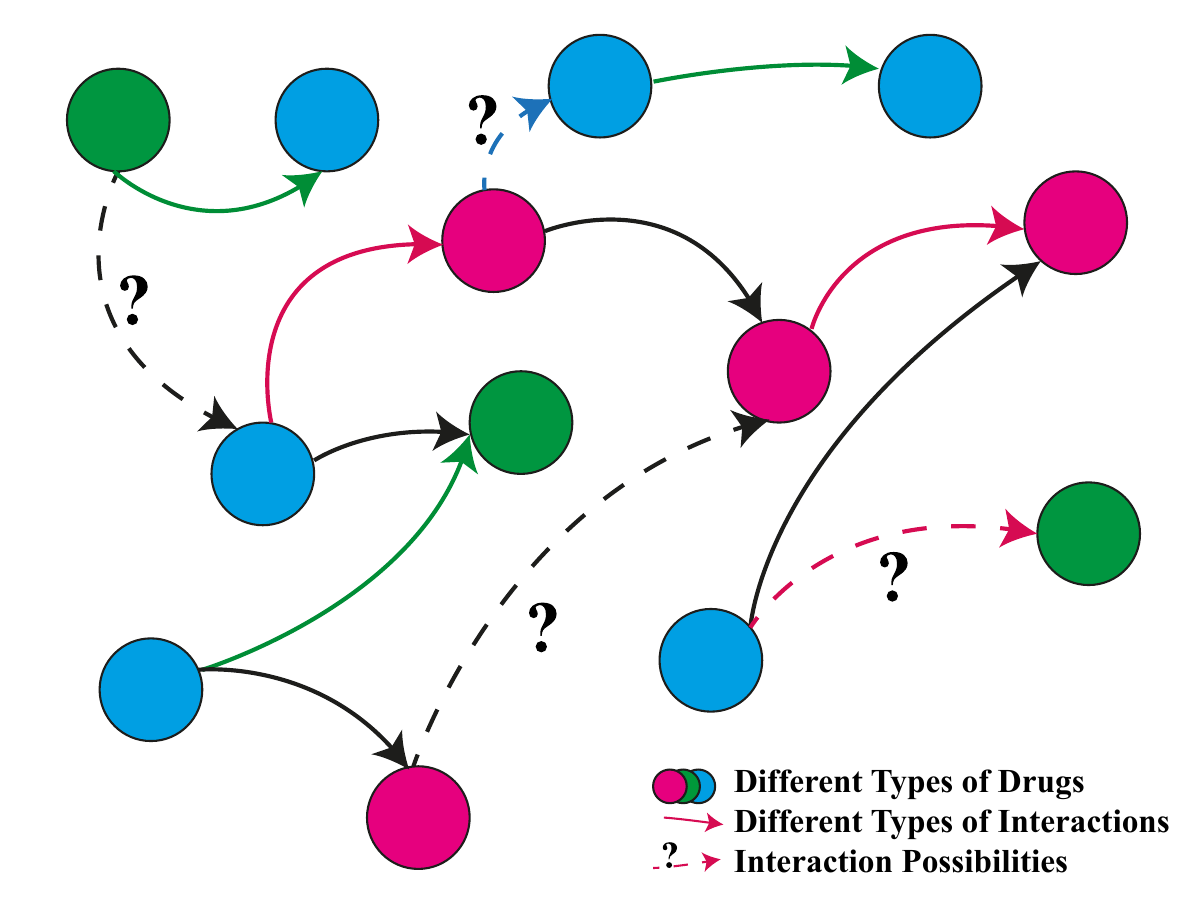}}
\caption{Drug-Drug Interactions Prediction Scenario}\label{fig:WHAT}
\end{figure}

Drug-Drug Interactions (DDIs) study is a key component of the drug development process.  When multiple medications are taken at the same time, complicated interactions between them can occur (illustrated in Figure \ref{fig:WHAT}). These interactions can provide important information about potential side effects and safety issues that can have a big impact on the creation and refinement of novel pharmacological agents. DDIs signify that when more than one or several drugs are taken simultaneously by patients, the potency of such medications may produce therapeutic effects or cause adverse consequences to arise. Unfortunately, identifying and predicting DDIs is challenging due to the vast number of potential drug interactions - making clinical examination of all combinations unfeasible, and resource-intensive. 

However, recent advancements in Graph Neural Networks (GNNs) suggest promising potential for DDI prediction. Originally, GNNs are renowned for their capability to represent nodes (like drugs in a drug-drug interaction network) and edges (like relationships between drugs) in a low-dimensional latent space that captures the subtleties of their interactions due to their topological and semantic characterization properties \cite{zitnik2018modeling, yu2021sumgnn, lin2020kgnn}. Notable GNNs include Graph Convolutional Networks (GCN) \cite{kipf2017semisupervised}, Graph Attention Networks (GAT) \cite{brody2022attentive}, and Graph Isomorphism Networks (GIN) \cite{xu2019powerful}. Owing to the success of GNNs, there have been remarkable studies on DDI prediction based on GNNs: \textbf{MSAN} \cite{10.1145/3511808.3557648} uses a Transformer-based GNN architecture; \textbf{MHCA-DDI} \cite{deac2019drugdrug} focuses on polypharmacy side effects; \textbf{GMPNN} \cite{10.1093/bib/bbab441} utilizes message-passing neural networks and a co-attention mechanism for predicting pairwise substructure interactions; \textbf{SSI-DDI} \cite{10.1093/bib/bbab133} employs substructure-substructure interaction modeling using Transformers; Ngo et al. \cite{ngo2022predicting} utilize variational graph autoencoder (\textbf{VGAE}) to capture polypharmacy side effects.




Despite the recent advances in DDI prediction, certain issues still remain unsolved and overlooked in this domain. \textbf{Firstly}, lack of generalization is a common weakness among the models discussed above. Most models face challenges when it comes to adapting to new graph structures or changes in node feature distribution. Relying heavily on molecular structure, MHCA-DDI \cite{deac2019drugdrug} lacks flexibility for generalization. SSI-DDI \cite{10.1093/bib/bbab133} faces limitations in terms of performance being affected by changes in drug order, while GMPNN \cite{10.1093/bib/bbab441} may struggle to capture complex mechanisms beyond substructures. 
\textbf{Secondly}, {feature extraction plays a crucial role in tackling complex tasks like DDI prediction. In this scenario, there are more than 1700 nodes representing drugs and 86 different types of interactions, resulting in millions of possible combinations. Furthermore, the challenge is intensified by the fact that any two drugs can have multiple interactions, and specific molecular structures of drug pairs contribute to distinct interactions. Take Ticlopidine (id DB00169) and Doxorubicin (id DB00997), for instance; their combination may both increase bioavailability (interaction type 5) and decrease serum concentration (interaction type 9) \cite{Wishart2018-ae}. To effectively capture these complex interaction, a robust feature extraction process is vital for any ML model to learn and predict novel DDIs, avoiding problems like overfitting, underfitting, biases, noises, etc.}
In terms of robust feature extraction, MSAN \cite{10.1145/3511808.3557648} attempts to overcome overfitting through similarity-based modules and substructure-dropping augmentation but struggles with feature extraction compared to graph autoencoders. SSI-DDI utilizes molecular graphs for segmentation and recognizes pairwise interactions, but can leak noisy information. GMPNN utilizes co-attention to learn drug substructures from molecular graphs, but struggles to capture intricate mechanisms beyond substructures, whereas GCN \cite{kipf2017semisupervised} falls short in distinguishing graphs with distinct edges and uniform nodes. 
{\textbf{Finally}, they all lack real-world representation and use capabilities. We argue that developing DDI models must be accompanied by proper real-world case studies analysis to check if previous clinical studies approve their applicability initially.} 


In order to address all aforementioned limitations, we propose \textbf{CADGL}, a novel \underline{C}ontext-\underline{A}ware \underline{D}eep \underline{G}raph \underline{L}earning framework. It is based on the VGAE architecture, but unlike the traditional VGAEs, our framework consists of two encoders: a context-aware graph encoder and a latent information encoder, and an MLP-based decoder. Our graph encoder aggregates node and edge values and positions across multiple layers to capture and process graph features from different perspectives, enabling robust feature extraction. Our approach has demonstrated greater effectiveness compared to these models in feature extraction and prediction, as evidenced by the experimental studies presented.

Our contributions are summarized into four folds: 
\begin{itemize}
    \item To the best of our knowledge, our work is the first attempt to exploit a context-aware framework based on our customized VGAE, incorporating both graph structural and latent information encoding to effectively capture the graph data for enhanced DDI prediction. 
    
    \item Our novel context-aware deep graph encoder involves two contextual pre-processors, namely Local and Molecular Context Pre-processors, to collect relations from different contexts and perspectives. It then encodes the acquired features via self-supervision, resulting in improved performance on downstream tasks and invariant representations that solve different issues posed by existing models.
    
    \item Experimental results show that CADGL outperforms all other state-of-the-art DDI prediction systems, with a detailed analysis of the strengths and necessity of different strategies adopted by our CADGL.

    \item Our model is able to predict valuable novel DDIs with strong clinical potential, supported by rigorous case studies confirming their applicability in drug development. We also analyze the broader societal implications of the drugs, which can be developed by newly identified DDIs, after comprehensive clinical experiments and studies.
\end{itemize}

\section{Related Works} \label{RelatedWorks}
\subsection{Drug-Drug Interactions} 
While DDIs are prevalent in real-world scenarios, there remains a scarcity of research addressing this issue. The proposed custom CNN \cite{MFConv} operates on graphs, enabling prediction streams using graphs of various sizes and shapes, extending molecular feature extraction beyond circular fingerprints. GraphDTA \cite{10.1093/bioinformatics/btaa921} aims to predict drug-target affinities, employing GNNs to create representations by constructing graphs for each molecule type. Lim et al. \cite{Lim2019PredictingDI} incorporate 3D structural data of protein-ligand binding poses into an adjacency matrix, differentiating inter-molecular relationships via gating and distance-aware graph attention. To predict DDIs, Torng and Altman \cite{Torng2018GraphCN} suggests a two-staged graph-convolutional network where an unsupervised graph-autoencoder is utilized in the initial step to learn fixed-size approximations of protein pockets.

\begin{figure*}[t] 
\centering {\includegraphics[scale=0.6]{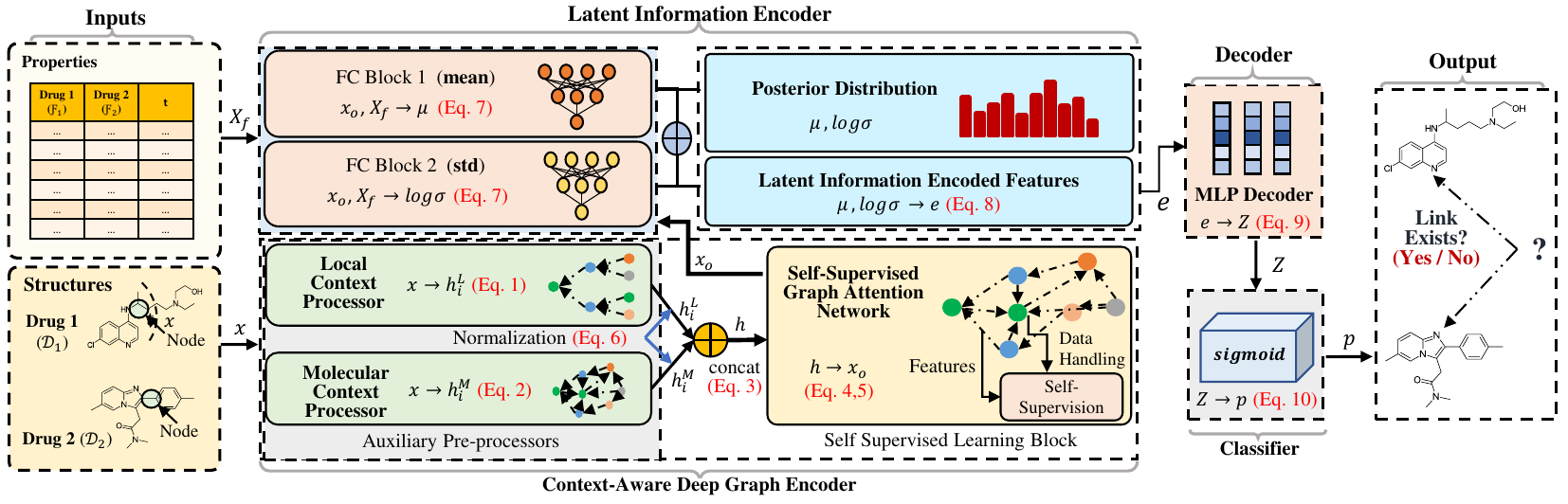}}
\caption{The overall framework of our CADGL. After taking input, structural data is processed with local and molecular context processors. These are passed into self-supervised graph attention for graph encoding. The output passes through two FC blocks with properties data for developing latent distribution and generate encodings. An MLP decoder extracts information, while a classifier gauges potential link likelihoods. }\label{fig:main}
\end{figure*}

\subsection{Context-Aware Learning in Drug-Interaction}
Context-Aware Learning is a research topic that explores how to use additional information (such as location, time, user preferences, etc.) to improve the performance of machine learning models.
Different contexts are usually used in Drug-Interaction\footnote{Alongside DDI, Drug-Interaction also covers Drug-Target Interaction, and Drug-Protein Interaction. These GNN tasks offer motivation in DDI research due to similarity in approaches.} related works, like molecular structure \cite{10.1145/3511808.3557648}, relations \cite{jiang2023relationaware}, etc. MSAN \cite{10.1145/3511808.3557648} proposes a novel model to effectively predict potential Drug-Drug Interactions (DDIs) using context of molecular structures of drug pairs. RaSECo \cite{jiang2023relationaware}  aims to enable all drugs, including new ones, to distill effective relation-aware embeddings (RaEs), thereby facilitating drug pair (DP) embedding to improve DDI prediction performance.

\subsection{Autoencoders in Drug-Interaction}
Variational Graph Autoencoder (VGAE) \cite{VGAE-main} is a framework for unsupervised graph learning where interpretable latent representations are developed using latent attributes for undirected graphs. To transform continuous code space vectors into tiny graphs by employing neural networks, GraphVAE \cite{Simonovsky2018GraphVAETG}, a molecule generation framework that consists of a decoder that outputs a probabilistic fully-connected graph which discards difficulty in chemical property representation. 

\section{Method and Algorithm}
Figure \ref{fig:main} illustrates the overview of our proposed CADGL framework. CADGL uses two encoders (graph and latent information) and an MLP-based decoder.
Given the structural inputs (shown in lower left corner of Figure \ref{fig:main}), we derive feature vectors using the context-aware deep graph encoder; subsequently, the latent information encoder refines these vectors to produce a latent information encoded feature vector, which is then combined with other inputs of physical and chemical properties. 
Finally, we feed it into the decoder, which generates a final prediction. 

\subsection{Problem Formulation}
GNNs utilize various message passing mechanisms to aggregate information from a center node's neighborhood and update its representation, using different techniques focusing on capturing different sets of features. Let $\mathcal{G}$ be a graph represented as $\mathcal{G}=(\mathcal{V}, \mathcal{E}, \mathcal{X})$, where $\mathcal{V}$ represents the nodes, $\mathcal{E}$ denotes the edges, and $\mathcal{X}$ is a matrix expressing the node features. 
Each node $v \in \mathcal{V}$ and each edge $e \in \mathcal{E}$ are associated with their mapping function $\phi(v): \mathcal{V} \rightarrow \mathcal{A}$ and $\varphi(e): \mathcal{E} \rightarrow \mathcal{R} . \mathcal{A}$ and $\mathcal{R}$ denote the node type set and edge type set, respectively, where $|\mathcal{A}|+|\mathcal{R}|>2$. 
The DDI prediction task involves binary classification using two tuples: $\left(\mathcal{D}_1, \mathcal{D}_2, t\right)$ and $\left({\mathcal{F}_1}, {\mathcal{F}_2}, t\right)$, where $\mathcal{D}$ denotes a drug structural data (chemical molecular graphs) obtained from $\mathcal{G}$ and ${\mathcal{F}}$ stands for physio-chemical properties of the drugs that come from $\mathcal{X}$, which is originated from RDKit \cite{greg_landrum_2023_7880616} using the DrugBank \cite{Wishart2018-ae} dataset. The goal is to predict if two drugs $\mathcal{D}_1, \mathcal{D}_2$ will interact under a specific interaction type $t$. 



\subsection{Context-Aware Deep Graph Encoder} 
Context-aware deep graph encoder encodes different contextual information effectively.
Our context-aware deep graph encoder (the lower area containing the three green blocks in Figure \ref{fig:main}), which is responsible for robust extraction of molecular structural information and generating feature vectors, comprises of two auxiliary information pre-processing units: {local context processor} (LCP) and {molecular context processor} (MCP), and a {self-supervised graph attention network}. LCP and MCP extract local structural and molecular characteristics and help the self-supervised network with relevant information to work effectively.

\subsubsection{Local context processor (LCP).} 
This module is responsible for processing local neighbourhood information using a graph convolutional neural network that utilizes a weight-sharing mechanism to enable efficient learning of node embeddings from the input graphs. It transforms each node's representation by aggregating information from its neighboring nodes and itself. 

To begin with, we make a structural input $x$ (see lower left corner of Figure \ref{fig:main}) by using a function $f_{graph}: \left(\mathcal{D}_1, \mathcal{D}_2, t\right) \rightarrow x$ that maps the given tuple to the input vectors, based on the concatenation and union operations on node features. Then the formulation of this layer is:
\begin{equation}
{h}_{i}^{L}={W} {x}_i +{W} \cdot \operatorname{MEAN}({x}_j, \forall j \in \mathcal{N}(i))
\end{equation}
where ${W}$ is a learnable parameter matrix for this layer and the MEAN operation aggregates the neighbor nodes ($\mathcal{N}(i)$) of $i$. The output ${h}_{i}^{L}$ is sent to the self supervised learning block as input.

\subsubsection{Molecular context processor (MCP).}
This processor leverages standard molecular feature extraction methods based on circular fingerprints. Drug molecules often contains circular structures like benzene and benzene-based substances, which play a vital role forming chemical bond and connections. MCP collects and processes these information for better prediction. Its formulation is:

\begin{equation}
{h}_{i}^{M}={W}_1^{(deg(i))}{x}_i+{W}_2^{(deg(i))} \sum_{j \in \mathcal{N}(i)} \mathbf{x}_j
\end{equation}
where ${W}_{1,2}$ are model parameters for this layer, and ${h}_i^{M}$ is the output. $deg$ denotes degree of a node, which is the count of its neighboring nodes to which it is directly connected by a edge. This process trains a unique weight matrix for each potential vertex degree in the molecular graph.

The outputs from LCP and MCP are concatenated to ${h}$:
\begin{equation}
{h} = concat ({h}_{i}^{L}, {h}_{i}^{M})
\end{equation}

\subsubsection{Self-supervised graph attention network (SSGAttn).} 
Self-supervised learning approaches for training GNNs have effective learning capabilities \cite{xie2022selfsupervised}. This network enhances model performance by incorporating a self-supervised learning objective, which encourages the encoder to learn representations that are invariant to graph transformations. This layer augments the models' ability to generalize to unseen data, making it more robust and effective. The formulation is:
\begin{equation}
{x}_o=\alpha_{ii} W_{s} {{h}_i}+\sum_{j \in \mathcal{N}(i)} \alpha_{ij} W_{s} {{h}_j}
\end{equation}
Here, ${h}_{.}$ is the input achieved from the pre-processors after concatenation, ${W}_s$ is a learnable parameter matrix, and ${x}_o$ is the output of the entire graph encoder network. $\alpha$ is related to the attention type. In this layer, we specially use the \textit{max attention}, namely $\alpha^{\text{MX}}$, computed by:

\begin{equation}
\begin{aligned}
\alpha^{\mathrm{MX}}_{ij} &=\frac{\exp \left(\operatorname{Leaky} \operatorname{ReLU}\left(e_{ij}^{\mathrm{MX}}\right)\right)}{\sum_{k \in \mathcal{N}(i) \cup\{i\}} \exp \left(\operatorname{LeakyReLU}\left(e_{ik}^{\mathrm{MX}}\right)\right)} \\
e^{\mathrm{MX}}_{ij} &= {a}^{\top}\left[W_{s} {{h}_i} \| W_{s} {{h}_j}\right] \cdot \sigma\left(\left(W_{s} {{h}_i}\right)^{\top}\cdot W_{s} {{h}_j}\right)
\end{aligned}
\end{equation}

{The self-supervision loss is applied to this layer, which uses positive and negative edge sampling. The loss term is described in Section \ref{sec:ssloss}.}

\subsubsection{Graph normalization layer.} 
Graph normalization layers are used to normalize output values from convolutional layers using GraphNorm \cite{cai2021graphnorm} before proceeding to the next layer. Its general equation is:

\begin{equation}
{x}_{norm}=\frac{{x}-\zeta \odot \mathbb{E}[{x}]}{\sqrt{\operatorname{Var}[{x}-\zeta \odot \mathbb{E}[{x}]]+\epsilon}} \odot \gamma+\beta
\end{equation}
where $\zeta, \beta, \text{ and } \gamma$ are learnable parameters, $\epsilon$ is a hyperparameter to control stability, and $\mathbb{E}[]$ and $\operatorname{Var}$ denote the expectation and variation. For our graph normalization after the LCP, it takes ${h}_{i}^{L}$ as input $x$ and the output ${x}_{norm}$ is renamed to ${h}_{i}^{L}$ and sent to the concat layer (Eq.(3)). Respectively, after the MCP, it takes ${h}_{i}^{M}$ as input $x$ and the output ${x}_{norm}$ is renamed to ${h}_{i}^{M}$ and sent to the concat layer.

\subsection{Latent Information Encoder and MLP Decoder}
Traditional Graph AEs lack the ability to generate predictions for links on new and unknown high-level graphs \cite{ngo2022predicting} due to the complexity and high-dimensional nature of data. But our model employs a latent information encoder to extract and combine features (substructure identification, atom centrality, ECFP6, aromaticity modeling, atom classification, physiochemical parameters, chemical feature categories, and other core chemistry attributes) from the RDKit library \cite{greg_landrum_2023_7880616}.

To begin with, we use another mapping function $f_{feature}: \left({\mathcal{F}_1}, {\mathcal{F}_2}, t\right) \rightarrow X_f$ that maps the property information to the input vectors. Then, along with the molecular structural encoded data ${x}_o$, the property information $X_f$ can be represented into lower-dimensional latent space. This helps in reducing the complexity and sparsity of data. In addition, this enables us to extract relevant latent variables through two fully connected layers that provide mean $\mu$ and log of variance $\log\sigma$ to create a low-dimensional multivariate normal distribution. We calculate $\mu$ and l$\log\sigma$ by:\label{generative}
\begin{equation}
\begin{aligned}
\mu= W_{\mu}[\Psi(x_o) \mathbin\Vert \Psi(X_f)] , 
\log\sigma= W_{\sigma}[\Psi(x_o) \mathbin\Vert \Psi(X_f)]
\end{aligned}    
\end{equation}
where $W_{\mu}$ and $W_{\sigma}$ are the model parameter matrices, $\Psi$ signifies the normalization, and $\left[\hspace{0.3cm} \parallel \hspace{0.3cm} \right]$ indicates concatenation. $\mu$ and $\log \sigma$ are the mean and log of variance vectors (length of 64), respectively.

Now, our model is able to produce new node links and interaction possibilities on intricate graphs at a high level of complexity by condensing all features into a low dimensional latent space using $\mu$ and $\log\sigma$ developing a posterior distribution:
\begin{equation}
\begin{aligned}
e=\mu+\kappa(\log\sigma) \cdot \exp(0.5\cdot \log\sigma)    
\end{aligned}
\end{equation}
where $e$ is the output, reparameterized (generating non-uniform random numbers by transforming a base distribution) from $\mu$ and $\log\sigma$. $\kappa$ is a function that generates a normal distribution (
$\mathcal{N}(0,1)$) arbitrarily.

Finally, our decoder incorporates a series of MLP layers to compute possible edges between corresponding nodes using the output generated by the encoders: 
\begin{equation} \label{eq:Z}
\begin{aligned}
Z = \text{MLP}({e})
\end{aligned}
\end{equation}
where ${e}$ is the input and $Z$ is the output vectors.
Then, the final prediction (i.e., an edge value) on whether an interaction will occur or not is performed by: 
\begin{equation}
\small p = \sigma(z) , 
z = \text{FCL} (Z)
\end{equation}
where $z$ is a scalar value produced by a fully connected layer (FCL) and $p$ is the final output probability, ranging from 0 to 1. $\sigma(.)$ denotes the sigmoid function.

\begin{algorithm}[ht]
\caption{CADGL}\label{algo1}
\begin{algorithmic}[1]
\STATE\textbf{Input}: $(\mathcal{D}_1, \mathcal{D}_2, t)$ for each combination 
\STATE \textbf{Output}: $p$
\STATE Map structural information $f_{graph}: \left(\mathcal{D}_1, \mathcal{D}_2, t\right) \rightarrow x$
\STATE Map physio-chemical features $f_{features} \rightarrow X_f$
\FOR{DDI combinations in $x, X_f$ }
\STATE Local Context Processor: $x \rightarrow {h}_{i}^{L}$
\STATE Molecular Context Processor: $x \rightarrow {h}_{i}^{M}$
\STATE Concatenate both: $({h}_{i}^{L}, {h}_{i}^{M}) \rightarrow {h}$
\STATE  Self-Supervised Graph Attention Net: ${h}\rightarrow x_o$
\STATE  Latent Encoding: $X_f, x_o \rightarrow \mu, \log\sigma$
\STATE  Reparameterization: $\mu, \log\sigma \rightarrow e $
\STATE  Decoding: $e \rightarrow Z $
\STATE  Probability calculation: $Z \rightarrow p $
\ENDFOR
\end{algorithmic}
\end{algorithm}

\subsection{Training Objectives}
For training our CADGL, we employ three loss functions, one is based on the cross-entropy (CE), one is for the self-supervision layer (SS), and the other is based on Kullback-Leibler Divergence (KL). The final loss function is a linear combination of the three loss terms without using any hyper-parameter, in order to avoid the complex hyper-parameter tuning. 
\begin{equation}
l_{CADGL}=l_{CE} + l_{SS} + l_{KL}
\end{equation}
Note that the impact of each loss function is examined and reported in supplementary materials.

\subsubsection{Cross-entropy loss (CE Loss)} 
CE loss is designed to minimize the difference between the predicted probabilities of the positive edges and 1, and the predicted probabilities of the negative edges and 0, as follows: 
\begin{equation}
\small
l_{CE}=-\frac{1}{n} \sum_{i=1}^n\left(p_i^{\prime} \log \sigma\left(p_i\right)+\left(1-p_i^{\prime}\right) \log \left(1-\sigma\left(p_i\right)\right)\right)
\end{equation}
Here, $p_i$ is the output of the $i$-th drug pair (in equation \ref{eq:Z}), where $i$ ranges from 1 to $n$. The $\sigma(\cdot)$ denotes sigmoid function and label of the $i$-th drug pair is denoted by $p_i^{\prime}$.

\subsubsection{KL-divergence loss (KL Loss)} 
This loss is for our latent information encoder (Eq. (8)). The KL-divergence loss \cite{günder2022kullbackleiblerdivergence} is a measure of the difference between two probability distributions. Here, it measures the difference between the learned distribution (with parameters $\mu$ and log $\sigma$) and a unit Gaussian. It is used to stabilize the distribution by keeping a stable Gaussian shape, as follows:
\begin{equation}
l_{KL}={KL}(q(z|x)\|p(z)) = \frac{1}{2}\sum_{j=1}^{D} (1 + 2\log \sigma_j - \mu_j^2 - \sigma_j^2)
\end{equation}
here $q(z|x)$ is the learned distribution with $\mu$ and $\log\sigma$, $D$ denotes dimension and $p(z)$ is the unit Gaussian distribution with mean 0 and variance 1. 

\subsubsection{Self-Supervision loss (SS Loss)} \label{sec:ssloss}
{
It is for the self-supervised graph encoder. We define the self-supervised optimization objective as a binary cross-entropy loss, following \cite{kim2021how}:
\begin{equation}
l_{SS}=-\frac{1}{\left|\mathbb{E} \cup \mathbb{E}^{-}\right|} \sum_{(j, i) \in \mathbb{E} \cup \mathbb{E}^{-}} I(i,j) \cdot \log \phi_{i j}^l+I(i,j) \cdot \log \left(1-\phi_{i j}^l\right)
\end{equation}
where $I(i,j)$ is an indicator function that outputs 1 if $i==j$ and 0 otherwise. We use a subset of $\mathbb{E} \cup \mathbb{E}^{-}$ (positive and negative sampling of edges) sampled by probability $p_e \in(0,1]$ at each training iteration for a regularization effect from randomness. $\phi_{i j}^l$ denotes the probability
that an edge exists between $i$ and $j$ nodes.}

\section{Implementation and Experiments}
\subsection{Experimental Settings}
We use the \textbf{DrugBank} \cite{Wishart2018-ae} dataset for our experiments. This dataset includes 1,703 drugs and a total of 191,870 drug pairs spanning 86 DDI types. It also contains structural and chemical information about the drugs. 
To conduct our analysis, we segregated the dataset into three subsets: a training set comprising 115,185 drug pairs, a validation set containing 38,348 drug pairs, and a test set consisting of 38,337 drug pairs. We perform all the experiments 05 times to collect experimental results. We train our model for a total of 300 iterations with the learning rate of 0.001. 


\subsection{Comparisons with State-of-the-arts}
In this experiment, we employ the following baselines and state-of-the-arts: \textbf{GCN} \cite{kipf2017semisupervised}, \textbf{GAT} \cite{brody2022attentive}, \textbf{GIN} \cite{xu2019powerful}, \textbf{SSI-DDI} \cite{10.1093/bib/bbab133}, \textbf{MSAN} \cite{10.1145/3511808.3557648},\textbf{ MHCA-DDI} \cite{deac2019drugdrug}, \textbf{DSN-DDI}\cite{DSN-DDI}, \textbf{SA-DDI} \cite{sgaessgaweg34asg}, \textbf{VGAE} \cite{ngo2022predicting} and \textbf{GMPNN} \cite{10.1093/bib/bbab441}, to compare performance against \textbf{CADGL}. Table \ref{table:model_performance} reports the comparison results. 
CADGL performs the best among all competitors. Some models discussed in related works are not designed for DDI tasks, meanwhile some of the codebases are missing, so we do not manage to compare with them. Within the context of medical research, even slight improvement in metrics can wield the power to determine matters of life and death. Assessing by performance, we believe that our model holds great significance and possesses the potential to yield meaningful contributions. 

\begin{figure*}[h] 
\centering {\includegraphics[scale=.45 ]{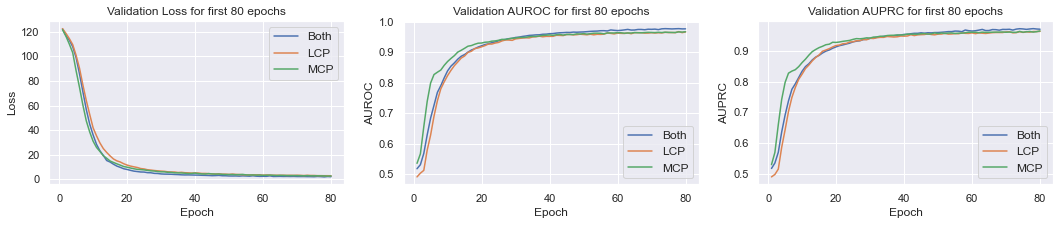}}
\vspace{-3mm}
\caption{Validation Loss, AUROC and AUPRC score visualization of context pre-processors for first 80 epochs.}\label{fig:AUC-ROC-CP}
\vspace{-3mm}
\end{figure*}
\begin{figure*}[t!] 
\centering {\includegraphics[scale=.99 ]{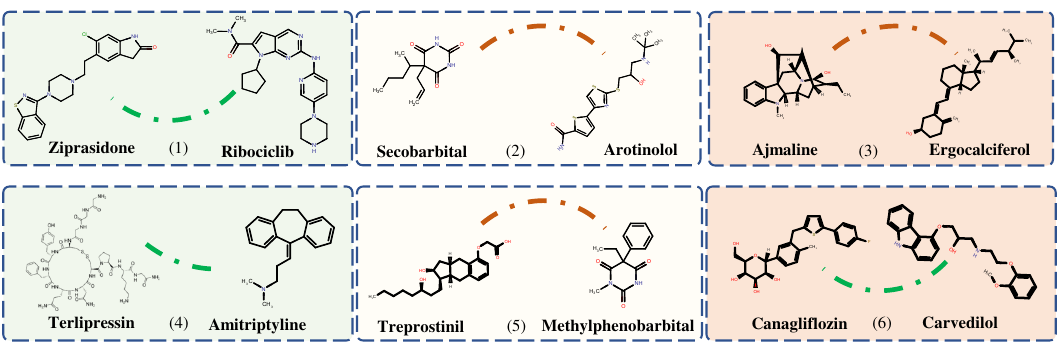}}
\vspace{-3mm}
\caption{Case study of new DDIs: (1) Ziprasidone and Ribociclib, (2) Secobarbital and Arotinolol,  (3) Ajmaline and Ergocalciferol, (4) Terlipressin and Amitriptyline,  (5) Treprostinil and Methylphenobarbital, and (3) Canagliflozin and Carvedilol.}\label{fig:case-study}
\vspace{-3mm}
\end{figure*}

\begin{table}[h]
\caption{Experimental results ($\uparrow$) (mean$\pm$std).}
\vspace{-1mm}
    \label{table:model_performance}
    \centering
    \begin{tabular}{c c c c c}
        \hline
        \textbf{Model} & \textbf{Accuracy (\%)} & \textbf{AUROC (\%)} & \textbf{F1 (x100)}\\
        
        \hline
        GCN (ICLR'17) & 81.42 $\pm$ 0.19 &  89.67 $\pm$ 0.14 & 82.94 $\pm$ 0.23 \\
        
        GIN (ICLR'19) & 96.08 $\pm$ 0.20 & 98.99 $\pm$ 0.04 & 96.18 $\pm$ 0.19 \\
        
        GAT (ICLR'22) & 94.00 $\pm$ 0.12  & 98.13 $\pm$ 0.05 & 94.22 $\pm$ 0.12 \\
        \hline
        
        MHCA (ICML-W'19) & 78.51 $\pm$ 0.15 &  86.33 $\pm$ 0.28 & 83.31 $\pm$ 0.21 \\
        
        SSI-DDI (BiB'19) & 96.33 $\pm$ 0.09 &  98.95 $\pm$ 0.08 & 96.38 $\pm$ 0.09 \\
        
        GMPNN (BiB'22) & 95.30 $\pm$ 0.05 &  98.46 $\pm$ 0.01 & 95.39 $\pm$ 0.05 \\
        
        VGAE (NIPS-W'22) & 92.12 $\pm$ 0.09  & 91.62 $\pm$ 0.14 & 89.12 $\pm$ 0.08  \\

        SA-DDI (Ch.Sci.'22)  & 96.23 $\pm$ 0.13  & 98.80 $\pm$ 0.08 & 96.29 $\pm$ 0.12  \\
        
        MSAN (CIKM'22) & 97.00 $\pm$ 0.09 &  99.27 $\pm$ 0.03 & 97.04 $\pm$ 0.08 \\
        
        DSN-DDI (BiB'23) & 96.94 $\pm$ 0.12 &  99.47 $\pm$ 0.08 & 96.93 $\pm$ 0.11 \\
        \hline
        
        \textbf{Ours (CADGL)}  & \textbf{98.21 $\pm$ 0.17} & \textbf{99.49 $\pm$ 0.14} & \textbf{97.79 $\pm$ 0.16} \\
        \hline
    \end{tabular}
    
\vspace{-3mm}
    
\end{table}

\subsection{Ablation Studies}

\subsubsection{Effectiveness of context-aware deep graph encoder}
Here in Table \ref{table:encoder_performance}, the study is conducted on our graph encoder and several baseline graph encoders. We change only the graph encoder by baselines. The latent information encoder and the decoder stay the same as our model in this ablation study. The study finds that our graph encoder outperforms all other alternatives, owing to its ability to aggregate node features by capturing local information and improve acquired features via self-supervision. 

\begin{table}[h]

\vspace{-3mm}
\caption{Comparison of different graph encoders.} \label{table:encoder_performance}
\vspace{-1mm}
    \centering
    \begin{tabular}{c c c c }
        \hline
        \textbf{Encoder} & \textbf{Accuracy (\%)} & \textbf{AUROC} & \textbf{F1 Score}\\
        \hline
        GCN & 92.12 $\pm$ 0.09  & 91.62 $\pm$ 0.14 & 89.12 $\pm$ 0.08 \\ 
        GAT  & 94.14 $\pm$ 0.12  & 94.53 $\pm$ 0.05 & 92.22 $\pm$ 0.14 \\ 
        SAGE  & 94.83 $\pm$ 0.31 & 95.68 $\pm$ 0.24 & 93.78 $\pm$ 0.11 \\ 
        GIN  & 94.51 $\pm$ 0.13  & 95.74 $\pm$ 0.32 & 94.01 $\pm$ 0.29 \\  
         
        \textbf{Ours} & \textbf{98.21 $\pm$ 0.17} & \textbf{99.49 $\pm$ 0.14} & \textbf{97.79 $\pm$ 0.16} \\
        \hline
        
\vspace{-3mm}
    \end{tabular}
\end{table}


\subsubsection{Impact of context pre-processors.} \label{sec:CPeffectiveness}
The ablation study conducted in Table \ref{tab:pre-processors} highlights the significance of both Local and Molecular Context Pre-processors (Local CP and Molecular CP) in achieving enhanced model performance. MCP has a more notable impact on the scores. The omission of either pre-processor notably reduces the model's performance, whereas the use of both enhances the scores. 

\begin{table}[h]
\vspace{-3mm}
\caption{Effectiveness of each pre-processors.}
\label{tab:pre-processors}
\vspace{-1mm}
\centering
\begin{tabular}{ c c c }
\hline
\textbf{Local CP} & \textbf{Molecular CP}  & \textbf{Accuracy (\%)} \\
\hline
\ding{55} & \ding{55}             &  -  \\ 
\ding{51} & \ding{55}             &  91.910 $\pm$ 0.236\%            \\ 
\ding{55} & \ding{51}             &  93.662 $\pm$ 0.152\%            \\  
\ding{51} & \ding{51}             &  \textbf{98.211 $\pm$ 0.171}\%    \\ 
\hline
\vspace{-3mm}
\end{tabular}
\end{table}

{Figure \ref{fig:AUC-ROC-CP} illustrates the validation loss, AUROC and AUPRC scores during the initial 80 epochs across various configurations. The results indicate that the model employing MCP shows a rapid initial learning phase, albeit reaching stabilization swiftly. Conversely, the combination of both LCP and MCP demonstrates a comparatively slower learning pace, accompanied by more pronounced improvements over time. }


\subsubsection{Effectiveness of our VGAE} \label{sec:VGAEeffectiveness}
Table \ref{table:model_performance} shows baseline model evaluations without a VGAE, while Table \ref{table:encoder_performance} presents results with VGAE. Adding a latent information encoder and decoder to GCN, which uses semi-supervised feature extraction, significantly boosts accuracy from 81.42\% to 92.12\%. However, GIN encoder's generative feature is ineffective due to fully isolating test and validation nodes and edges, hindering precise mapping to the latent space. GATs prioritize expressive node representations, but in a generative setting, our graph encoder with LCP, MCP, and SSGAttn components achieves the most optimal performance.

\section{Experimental Case Studies} \label{section:case study}
For real-world impact, a DDI prediction model should be able to predict novel undiscovered DDI in a very effective and productive manner, so that the drug development process can be accelerated. 

In Table \ref{tab:new drug}, we present some novel possible DDIs to illustrate our model's capability to accurately predict potential DDIs. The DrugBank ( \url{https://go.drugbank.com} and ECGWave website \url{https://ecgwaves.com/} ) is used to obtain evidence for the DDIs and verify. The table also displays a set of potential DDIs that are very likely to occur but have not been clinically identified yet. We also present a clinical research analysis of the undiscovered DDIs, exploring their possibility of occurring clinically in real life. As they are still undiscovered, we will explore related studies to determine if our model's predictions align with real-world situations. More examples are included in the appendix.

\subsection{Clinical Study Analysis on Novel DDIs} \label{cilinical-analyss}
Here, we analyze clinical studies of some newly predicted DDIs for drug development {(visualized in Figure \ref{fig:case-study})} identified by our model for potential clinical application in order to validate predictions and model performance:


  \textit{1. QTc-prolonging activities of Ribociclib increases by Ziprasidone}. QTc-prolonging happens when it takes longer than usual for the heart to recharge between beats \cite{10.1093/toxsci/kfac013}. Ziprasidone can effectively reduce the rate and time of relapses in schizophrenia and it prolongs QTc interval and may block potassium channels in cardiac cells \cite{Wishart2018-ae}. Combination of Ziprasidone with Ribociclib can increase the risk of QTc prolongation \cite{Taylor2003-jh} as predicted by our model with 90.109\% confidence, implicating probability of creating a novel drug which can be beneficial in this case.
 
  \textit{2. Hypotensive activities of Arotinolol increases by Secobarbital}.   Hypotension is a decrease in systemic blood pressure below accepted low values. While there is not an accepted standard hypotensive value, pressures less than 90/60 are recognized as hypotensive \cite{Sharma_Hashmi_Bhattacharya_2023}. Our model suggests (with 99.883\% confidence) that combining Arotinolol, which is being studied for its potential as antihypertensive therapy \cite{Wu2001-qp}, with Secobarbital having anaesthetic, anticonvulsant, sedative, and hypnotic properties \cite{Wishart2018-ae} may result in increased efficacy in hypotensive activities.

  \textit{3. Arrhythmogenic fixation activities of Ajmaline increases by Ergocalciferol}. Arrhythmogenic activities means the ability of a substance or condition to cause arrhythmias or abnormal heart rhythms. Ajmaline, a class 1A antiarrhythmic, enhances heart rhythm by modulating sodium channels \cite{Brugada2003}, fixing arrhythmogenic problems. Ergocalciferol, a form of vitamin D, has been shown to have a positive impact on cardiovascular health. \cite{LipidsonCardiovascular} So, the likelihood of intensifying arrhythmogenic fixation effects is noteworthy, as noted by our model.

  \textit{4. Antihypertensive activities of Terlipressin decreases by Amitriptyline}.  Antihypertensive activities of drugs refer to the ability of drugs to help lowering high blood pressure. Terlipressin's pharmacokinetic and molecular attributes confer some benefits like preventing rebound hypotension upon drug cessation \cite{Pesaturo2006}. Amitriptyline with analgesic properties, can lead to ECG changes, and increased glucose level \cite{Bryson1996}. Given contradictory outputs, our model suggestion of reduced antihypertension can be effective.

  \textit{5. Hypotensive activities of Treprostinil increases by Methylphenobarbital}. Treprostinil exhibits hypotensive effects by inhibiting pulmonary vasculature vasoconstriction \cite{hypotensivedrfff}. Methylphenobarbital is usually utilized for tension headaches and migraines, exerting sedative effects through CNS depression. Combining both substances can result in increased hypotensive effects as per our model.

  \textit{6. Hypoglycemic activities of Canagliflozin increases by Carvedilol}. Hypoglycemic activity refers to the ability of a drug to help lowering the blood sugar, often used for treating diabetes. Canagliflozin has been shown to improve glycemic control in patients with type 2 diabetes (T2D) \cite{Minami2021}. In a study on recurrently hypoglycemic rats, carvedilol was found to improve hypoglycemia awareness \cite{ratshypg}. So, there is a significant potential for Carvedilol to amplify the hypoglycemic effects of Canagliflozin, as suggested by our model.

\begin{table*}[bt!]
\caption{Potential new DDIs.} \label{tab:new drug}
\centering
\begin{tabular}{c c c c c}
\hline
\textbf{Serial Number} & \textbf{Drug 1 ($\mathcal{D}_1$)} & \textbf{Drug 2 ($\mathcal{D}_2$)} & \textbf{Interaction Type (t)} & \textbf{Probability of Occuring} \\
\hline

\multicolumn{5}{c}{\textbf{New Drug-Drug Interactions (Clinically Verified via DrugBank)}} \\ 
\hline
       1 & Verapamil & Pravastatin & Drug 2  Serum concentration increases ($\uparrow$) & 99.919\% \\ 
       2 &  Voriconazole & Apixaban & Drug 2  Metabolism decreases ($\downarrow$) & 99.624\%  \\ 
       3 &  Fentanyl & Vortioxetine & Risk of adverse effects increases ($\uparrow$) & 99.549\%  \\ 
       4 &  Vemurafenib & Rifaximin & Drug 2  Serum Concentration increases ($\uparrow$) & 99.849\%  \\ 
       5 &  Erythromycin & Fentanyl & Drug 2  Metabolism decreases ($\downarrow$) & 99.385\%  \\ 
\hline
\multicolumn{5}{c}{\textbf{{Possible Novel Drug-Drug Interactions (Not Clinically Verified Yet)}}} \\ 
\hline
      1 &   Secobarbital & Arotinolol & Drug 2  hypotensive activities increases ($\uparrow$) & 99.883\%  \\ 
      2 &   Mesalazine & Nafamostat & Drug 2  anticoagulant activities increases ($\uparrow$) & 99.833\%  \\ 
      3 &   Voriconazole & Etoposide & Drug 2  Serum Concentration increases ($\uparrow$) & 99.816\%  \\ 
     4 &   Risperidone & Propofol & Risk of adverse effects increases ($\uparrow$) & 99.660\%  \\ 
      5 &   Pipamperone & Bupivacaine & Risk of adverse effects increases ($\uparrow$) & 99.503\%  \\ 
       6 &  Moricizine & Hydralazine & Drug 2  hypotensive activities increases ($\uparrow$)  & 99.491\%  \\ 
      7 &   Isoniazid & Tazarotene & Drug 2  Metabolism decreases ($\downarrow$) & 99.420\%  \\ 
     8 &    Methylphenobarbital & Treprostinil & Drug 2  hypotensive activities increases ($\uparrow$)  & 99.366\% \\ 
      9 &   Nevirapine & Vinorelbine & Drug 2  Serum Concentration decreases ($\downarrow$) & 99.337\%  \\ 
      10 &   Dabrafenib & Pentamidine & Drug 2  QTc-prolongation increases ($\uparrow$)  & 98.813\%  \\
      \hline

\end{tabular}
\end{table*}

\section{Discussion}
Continuous new drug development is important for many reasons, such as treating new diseases, preventing the spread of infections, overcoming drug resistance, creating better drugs with less side-effects, and improving the quality of life for patients. However, developing new drugs is a complex and costly process that requires scientific innovation, rigorous testing, and regulatory approval \cite{Mohs2017}, which can be accelerated utilizing DDIs. Our approach has also demonstrated encouraging results in advancing new drug development using DDIs. Our model presents intriguing DDIs discussed through case studies involving CVD, hyper/hypotension, and hypoglycemic activities, showing significant potential for clinical testing and substantial impact.

It is also very important to clarify that our work does not guarantee the development of novel drugs, nor do we lay claim to the ability to create new pharmaceuticals directly. Instead, our focus lies on expediting the drug discovery process by producing effective drug-drug interaction combinations and gauging their potential efficacy. The primary goal is to provide valuable insights so that clinical researchers can streamline their efforts, avoiding exhaustive testing that consumes significant time and resources. Drug development is an intricate and multifaceted undertaking. Our contribution is positioned as a facilitative effort aimed at easing the burden on researchers by showing the possibilities of new drugs with the help of artificial intelligence. By identifying promising combinations and assessing their potential effectiveness, our research aims to empower clinical researchers by offering a valuable roadmap, enabling researchers to navigate the complexities of drug development more efficiently. Our goal is to contribute to the scientific community by reducing the time and effort expended on extensive pre-clinical testing, ultimately expediting the translation of discoveries into clinical trials.

\section{Conclusion}
In this paper, we introduce CADGL, a novel context-aware deep graph learning framework that endeavors to predict DDIs. Our approach includes a three-component variational graph autoencoder architecture with a graph encoder, a latent information encoder, and a decoder. The context-aware deep graph encoder captures node embeddings, the latent information encoder transforms embeddings and properties enabling generative modeling, and an MLP decoder predicts links between drug nodes. Effectiveness of CADGL is supported by compelling evidence of possible new drugs which can be developed by newly identified DDIs, and extensive experimentation confirms its superiority over various state-of-the-art models.



\section*{Acknowledgments}
This work was partly supported by (1) Institute of Information \& communications Technology Planning \& Evaluation (IITP) grant funded by the Korea government (MSIT) (No.2020-0-01373, Artificial Intelligence Graduate School Program (Hanyang University)) and (2) the National Research Foundation of Korea (NRF) grant funded by the Korea government (*MSIT) (No.2018R1A5A7059549). *Ministry of Science and ICT.



{\small
\bibliographystyle{IEEEtran}
\bibliography{our_paper}

@inproceedings{brody2022attentive,
      title={How Attentive are Graph Attention Networks?}, 
author={Shaked Brody and Uri Alon and Eran Yahav},
booktitle={International Conference on Learning Representations},
year={2022},
url={https://openreview.net/forum?id=F72ximsx7C1}
}

@misc{günder2022kullbackleiblerdivergence,
      title={Full Kullback-Leibler-Divergence Loss for Hyperparameter-free Label Distribution Learning}, 
      author={Maurice Günder and Nico Piatkowski and Christian Bauckhage},
      year={2022},
      eprint={2209.02055},
      archivePrefix={arXiv},
      primaryClass={cs.LG}
}

@article{zitnik2018modeling,
  title={Modeling polypharmacy side effects with graph convolutional networks},
  author={Zitnik, Marinka and Agrawal, Monica and Leskovec, Jure},
  journal={Bioinformatics},
  volume={34},
  number={13},
  pages={i457--i466},
  year={2018},
  publisher={Oxford University Press}
}

@ARTICLE{xie2022selfsupervised,
  author={Xie, Yaochen and Xu, Zhao and Zhang, Jingtun and Wang, Zhengyang and Ji, Shuiwang},
  journal={IEEE Transactions on Pattern Analysis and Machine Intelligence}, 
  title={Self-Supervised Learning of Graph Neural Networks: A Unified Review}, 
  year={2023},
  volume={45},
  number={2},
  pages={2412-2429},
  doi={10.1109/TPAMI.2022.3170559}}

@article{yu2021sumgnn,
  title={SumGNN: multi-typed drug interaction prediction via efficient knowledge graph summarization},
  author={Yu, Yue and Huang, Kexin and Zhang, Chao and Glass, Lucas M and Sun, Jimeng and Xiao, Cao},
  journal={Bioinformatics},
  volume={37},
  number={18},
  pages={2988--2995},
  year={2021},
  publisher={Oxford University Press}
}

@inproceedings{lin2020kgnn,
  title={KGNN: Knowledge Graph Neural Network for Drug-Drug Interaction Prediction.},
  author={Lin, Xuan and Quan, Zhe and Wang, Zhi-Jie and Ma, Tengfei and Zeng, Xiangxiang},
  booktitle={IJCAI},
  volume={380},
  pages={2739--2745},
  year={2020}
}

@inproceedings{kipf2017semisupervised,
      title={Semi-Supervised Classification with Graph Convolutional Networks}, 
author={Thomas N. Kipf and Max Welling},
booktitle={International Conference on Learning Representations},
year={2017},
url={https://openreview.net/forum?id=SJU4ayYgl}
}

@inproceedings{
kim2021how,
title={How to Find Your Friendly Neighborhood: Graph Attention Design with Self-Supervision},
author={Dongkwan Kim and Alice Oh},
booktitle={International Conference on Learning Representations},
year={2021},
url={https://openreview.net/forum?id=Wi5KUNlqWty}
}

@inproceedings{xu2019powerful,
      title={How Powerful are Graph Neural Networks?}, 
author={Keyulu Xu and Weihua Hu and Jure Leskovec and Stefanie Jegelka},
booktitle={International Conference on Learning Representations},
year={2019},
url={https://openreview.net/forum?id=ryGs6iA5Km},
}

@inproceedings{10.1145/3511808.3557648,
author = {Zhu, Xinyu and Shen, Yongliang and Lu, Weiming},
title = {Molecular Substructure-Aware Network for Drug-Drug Interaction Prediction},
year = {2022},
isbn = {9781450392365},
publisher = {Association for Computing Machinery},
address = {New York, NY, USA},
url = {https://doi.org/10.1145/3511808.3557648},
doi = {10.1145/3511808.3557648},
booktitle = {Proceedings of the 31st ACM International Conference on Information and Knowledge Management},
pages = {4757–4761},
numpages = {5},
location = {Atlanta, GA, USA},
series = {CIKM '22}
}

@InProceedings{cai2021graphnorm,
  title = 	 {GraphNorm: A Principled Approach to Accelerating Graph Neural Network Training},
  author =       {Cai, Tianle and Luo, Shengjie and Xu, Keyulu and He, Di and Liu, Tie-Yan and Wang, Liwei},
  booktitle = 	 {Proceedings of the 38th International Conference on Machine Learning},
  pages = 	 {1204--1215},
  year = 	 {2021},
  editor = 	 {Meila, Marina and Zhang, Tong},
  volume = 	 {139},
  series = 	 {Proceedings of Machine Learning Research},
  month = 	 {18--24 Jul},
  publisher =    {PMLR},
  url = 	 {https://proceedings.mlr.press/v139/cai21e.html}
}

@inproceedings{ngo2022predicting,
      title={Predicting Drug-Drug Interactions using Deep Generative Models on Graphs}, 
author={Khang Nhat Ngo and Truong Son Hy and Risi Kondor},
booktitle={NeurIPS 2022 AI for Science: Progress and Promises},
year={2022},
url={https://openreview.net/forum?id=Hnax-9OBNtH}
}

@ARTICLE{Wishart2018-ae,
  title    = "{DrugBank} 5.0: a major update to the {DrugBank} database for
              2018",
  author   = "Wishart, David S and Feunang, Yannick D and Guo, An C and Lo,
              Elvis J and Marcu, Ana and Grant, Jason R and Sajed, Tanvir and
              Johnson, Daniel and Li, Carin and Sayeeda, Zinat and Assempour,
              Nazanin and Iynkkaran, Ithayavani and Liu, Yifeng and
              Maciejewski, Adam and Gale, Nicola and Wilson, Alex and Chin,
              Lucy and Cummings, Ryan and Le, Diana and Pon, Allison and Knox,
              Craig and Wilson, Michael",
  journal  = "Nucleic Acids Res.",
  volume   =  46,
  number   = "D1",
  pages    = "D1074--D1082",
  month    =  jan,
  year     =  2018,
  language = "en"
}

@article{10.1093/bib/bbab133,
    author = {Nyamabo, Arnold K and Yu, Hui and Shi, Jian-Yu},
    title = "{SSI–DDI: substructure–substructure interactions for drug–drug interaction prediction}",
    journal = {Briefings in Bioinformatics},
    volume = {22},
    number = {6},
    year = {2021},
    month = {05},
    issn = {1477-4054},
    doi = {10.1093/bib/bbab133},
    url = {https://doi.org/10.1093/bib/bbab133},
    note = {bbab133},
    eprint = {https://academic.oup.com/bib/article-pdf/22/6/bbab133/41087527/supplementary-materials\_bbab133.pdf},
}

@article{10.1093/bib/bbab441,
    author = {Nyamabo, Arnold K and Yu, Hui and Liu, Zun and Shi, Jian-Yu},
    title = "{Drug–drug interaction prediction with learnable size-adaptive molecular substructures}",
    journal = {Briefings in Bioinformatics},
    volume = {23},
    number = {1},
    year = {2021},
    month = {10},
    issn = {1477-4054},
    doi = {10.1093/bib/bbab441},
    url = {https://doi.org/10.1093/bib/bbab441},
    note = {bbab441},
    eprint = {https://academic.oup.com/bib/article-pdf/23/1/bbab441/42231788/supplementary-materials\_bbab441.pdf},
}

@misc{deac2019drugdrug,
      title={Drug-Drug Adverse Effect Prediction with Graph Co-Attention}, 
      author={Andreea Deac and Yu-Hsiang Huang and Petar Veličković and Pietro Liò and Jian Tang},
      year={2019},
      eprint={1905.00534},
      archivePrefix={arXiv},
      primaryClass={stat.ML}
}

@ARTICLE{Wu2001-qp,
  title     = "Clinical trial of arotinolol in the treatment of hypertension:
               dippers vs. non-dippers",
  author    = "Wu, H and Zhang, Y and Huang, J and Zhang, Y and Liu, G and Sun,
               N and Yu, Z and Zhou, Y",
  journal   = "Hypertens. Res.",
  publisher = "Japanese Society of Hypertension",
  volume    =  24,
  number    =  5,
  pages     = "605--610",
  month     =  sep,
  year      =  2001,
  language  = "en"
}

@ARTICLE{Taylor2003-jh,
  title    = "Ziprasidone in the management of schizophrenia : the {QT}
              interval issue in context",
  author   = "Taylor, David",
  journal  = "CNS Drugs",
  volume   =  17,
  number   =  6,
  pages    = "423--430",
  year     =  2003,
  language = "en"
}

@ArtifactSoftware{R,
    title = {R: A Language and Environment for Statistical Computing},
    author = {{R Core Team}},
    organization = {R Foundation for Statistical Computing},
    address = {Vienna, Austria},
    year = {2019},
    url = {https://www.R-project.org/},
}

@software{greg_landrum_2023_7880616,
  author       = {Greg Landrum and
                  Paolo Tosco and
                  Brian Kelley and
                  Ric and
                  sriniker and
                  David Cosgrove and
                  gedeck and
                  Riccardo Vianello and
                  NadineSchneider and
                  Eisuke Kawashima and
                  Dan N and
                  Gareth Jones and
                  Andrew Dalke and
                  Brian Cole and
                  Matt Swain and
                  Samo Turk and
                  AlexanderSavelyev and
                  Alain Vaucher and
                  Maciej Wójcikowski and
                  Ichiru Take and
                  Daniel Probst and
                  Kazuya Ujihara and
                  Vincent F. Scalfani and
                  guillaume godin and
                  Axel Pahl and
                  Francois Berenger and
                  JLVarjo and
                  Rachel Walker and
                  jasondbiggs and
                  strets123},
  title        = {rdkit/rdkit: 2023\_03\_1 (Q1 2023) Release},
  month        = apr,
  year         = 2023,
  publisher    = {Zenodo},
  version      = {Release\_2023\_03\_1},
  doi          = {10.5281/zenodo.7880616},
  url          = {https://doi.org/10.5281/zenodo.7880616}
}

@article{10.1093/toxsci/kfac013,
    author = {Valentin, Jean-Pierre and Hoffmann, Peter and Ortemann-Renon, Catherine and Koerner, John and Pierson, Jennifer and Gintant, Gary and Willard, James and Garnett, Christine and Skinner, Matthew and Vargas, Hugo M and Wisialowski, Todd and Pugsley, Michael K},
    title = "{The Challenges of Predicting Drug-Induced QTc Prolongation in Humans}",
    journal = {Toxicological Sciences},
    volume = {187},
    number = {1},
    pages = {3-24},
    year = {2022},
    month = {02},
    issn = {1096-6080},
    doi = {10.1093/toxsci/kfac013},
    url = {https://doi.org/10.1093/toxsci/kfac013},
    eprint = {https://academic.oup.com/toxsci/article-pdf/187/1/3/43465751/kfac013.pdf},
}

@article{Sharma_Hashmi_Bhattacharya_2023, title={Hypotension - StatPearls - NCBI Bookshelf}, url={https://www.ncbi.nlm.nih.gov/books/NBK499961/}, journal={NCBI Bookshelf}, publisher={NCBI}, author={Sharma, Sandeep and Hashmi, Muhammad&nbsp; F. and Bhattacharya, Priyanka&nbsp; T.}, year={2023}, month={Feb}}

@article{Mohs2017,
    author = {Lin, Xuan and Dai, Lichang and Zhou, Yafang and Yu, Zu-Guo and Zhang, Wen and Shi, Jian-Yu and Cao, Dong-Sheng and Zeng, Li and Chen, Haowen and Song, Bosheng and Yu, Philip S and Zeng, Xiangxiang},
    title = "{Comprehensive evaluation of deep and graph learning on drug–drug interactions prediction}",
    journal = {Briefings in Bioinformatics},
    volume = {24},
    number = {4},
    pages = {bbad235},
    year = {2023},
    month = {07},
    issn = {1477-4054},
    doi = {10.1093/bib/bbad235}
}

@article{DSN-DDI,
    author = {Li, Zimeng and Zhu, Shichao and Shao, Bin and Zeng, Xiangxiang and Wang, Tong and Liu, Tie-Yan},
    title = "{DSN-DDI: an accurate and generalized framework for drug–drug interaction prediction by dual-view representation learning}",
    journal = {Briefings in Bioinformatics},
    volume = {24},
    number = {1},
    pages = {bbac597},
    year = {2023},
    month = {01},
    issn = {1477-4054},
    doi = {10.1093/bib/bbac597},
    url = {https://doi.org/10.1093/bib/bbac597},
    eprint = {https://academic.oup.com/bib/article-pdf/24/1/bbac597/48782745/bbac597.pdf},
}

@article{Minami2021,
  doi = {10.1080/14656566.2021.1939675},
  url = {https://doi.org/10.1080/14656566.2021.1939675},
  year = {2021},
  month = jun,
  publisher = {Informa {UK} Limited},
  volume = {22},
  number = {16},
  pages = {2087--2094},
  author = {Taichi Minami and Akiko Kameda and Yasuo Terauchi},
  title = {An evaluation of canagliflozin for the treatment of type 2 diabetes: an update},
  journal = {Expert Opinion on Pharmacotherapy}
}

@Journal{ ratshypg,
        title={ Carvedilol Prevents Counterregulatory Failure in Recurrently Hypoglycemic Rats },
        author={ Rawad Farhat and Gong Su and Owen Chan },
        journal={ Diabetes },
        year={ 2018 },
        doi={ 10.2337/DB18-200-OR },  
      }

@article{Brugada2003,
  doi = {10.1016/s0195-668x(03)00232-x},
  url = {https://doi.org/10.1016/s0195-668x(03)00232-x},
  year = {2003},
  month = jun,
  publisher = {Oxford University Press ({OUP})},
  volume = {24},
  number = {12},
  pages = {1085--1086},
  author = {J Brugada},
  title = {The ajmaline challenge in Brugada syndrome A useful tool or misleading information?},
  journal = {European Heart Journal}
}

@Journal { LipidsonCardiovascular,
        title={ Impact of Lipids on Cardiovascular Health: JACC Health Promotion Series },
        author={ Brian A. Ference and Ian D. Graham and Lale Tokgozoglu and Alberico L. Catapano },
        journal={ Journal of the American College of Cardiology },
        year={ 2018 },
        doi={ 10.1016/J.JACC.2018.06.046 },  
      }

@article{Pesaturo2006,
  doi = {10.1345/aph.1h373},
  url = {https://doi.org/10.1345/aph.1h373},
  year = {2006},
  month = dec,
  publisher = {{SAGE} Publications},
  volume = {40},
  number = {12},
  pages = {2170--2177},
  author = {Adam B Pesaturo and Heath R Jennings and Stacy A Voils},
  title = {Terlipressin: Vasopressin Analog and Novel Drug for Septic Shock},
  journal = {Annals of Pharmacotherapy}
}

@article{Bryson1996,
  doi = {10.2165/00002512-199608060-00008},
  year = {1996},
  month = jun,
  publisher = {Springer Science and Business Media {LLC}},
  volume = {8},
  number = {6},
  pages = {459--476},
  author = {Harriet M. Bryson and Michelle I. Wilde},
  title = {Amitriptyline},
  journal = {Drugs and Aging}
}

@Journal{ hypotensivedrfff,
        title={ Inhaled hexadecyl-treprostinil provides pulmonary vasodilator activity at significantly lower plasma concentrations than infused treprostinil. },
        author={ Richard W. Chapman and Zhili Li and Michel R. Corboz and Helena Gauani and Adam J. Plaunt and Donna M. Konicek and Franziska Leifer and Charles E. Laurent and Han Yin and Dany Salvail and Chad Dziak and Walter Perkins and Vladimir Malinin },
        journal={ Pulmonary Pharmacology & Therapeutics },
        year={ 2018 },
        doi={ 10.1016/J.PUPT.2018.02.002 },  
      }

@inproceedings{MFConv,
      title={Convolutional Networks on Graphs for Learning Molecular Fingerprints}, 
author = {Duvenaud, David and Maclaurin, Dougal and Aguilera-Iparraguirre, Jorge and G\'{o}mez-Bombarelli, Rafael and Hirzel, Timothy and Aspuru-Guzik, Al\'{a}n and Adams, Ryan P.},
year = {2015},
publisher = {MIT Press},
address = {Cambridge, MA, USA},
booktitle = {Proceedings of the 28th International Conference on Neural Information Processing Systems - Volume 2},
pages = {2224–2232},
numpages = {9},
location = {Montreal, Canada},
series = {NIPS'15}
}

@article{Lim2019PredictingDI,
  title={Predicting Drug-Target Interaction Using a Novel Graph Neural Network with 3D Structure-Embedded Graph Representation},
  author={Jaechang Lim and Seongok Ryu and Kyubyong Park and Yo Joong Choe and Jiyeon Ham and Woo Youn Kim},
  journal={Journal of chemical information and modeling},
  year={2019},
  url={https://api.semanticscholar.org/CorpusID:201631371}
}

@article{Torng2018GraphCN,
  title={Graph Convolutional Neural Networks for Predicting Drug-Target Interactions},
  author={Wen Torng and Russ B. Altman},
  journal={bioRxiv},
  year={2018},
  url={https://api.semanticscholar.org/CorpusID:92451162}
}

@misc{VGAE-main,
      title={Variational Graph Auto-Encoders}, 
      author={Thomas N. Kipf and Max Welling},
      year={2016},
      eprint={1611.07308},
      archivePrefix={arXiv},
      primaryClass={stat.ML}
}

@inproceedings{Simonovsky2018GraphVAETG,
  title={GraphVAE: Towards Generation of Small Graphs Using Variational Autoencoders},
  author={Martin Simonovsky and Nikos Komodakis},
  booktitle={International Conference on Artificial Neural Networks},
  year={2018},
  url={https://api.semanticscholar.org/CorpusID:3637466}
}

@article{10.1093/bioinformatics/btaa921,
    author = {Nguyen, Thin and Le, Hang and Quinn, Thomas P and Nguyen, Tri and Le, Thuc Duy and Venkatesh, Svetha},
    title = "{GraphDTA: predicting drug–target binding affinity with graph neural networks}",
    journal = {Bioinformatics},
    volume = {37},
    number = {8},
    pages = {1140-1147},
    year = {2020},
    month = {10},
    issn = {1367-4803},
    doi = {10.1093/bioinformatics/btaa921},
    url = {https://doi.org/10.1093/bioinformatics/btaa921},
    eprint = {https://academic.oup.com/bioinformatics/article-pdf/37/8/1140/50340643/btaa921.pdf},
}

@article{jiang2023relationaware,
title = {Relation-aware graph structure embedding with co-contrastive learning for drug–drug interaction prediction},
journal = {Neurocomputing},
volume = {572},
pages = {127203},
year = {2024},
issn = {0925-2312},
doi = {https://doi.org/10.1016/j.neucom.2023.127203},
}

@article{sgaessgaweg34asg,
  title = {Learning size-adaptive molecular substructures for explainable drug–drug interaction prediction by substructure-aware graph neural network},
  volume = {13},
  ISSN = {2041-6539},
  DOI = {10.1039/d2sc02023h},
  number = {29},
  journal = {Chemical Science},
  publisher = {Royal Society of Chemistry (RSC)},
  author = {Yang,  Ziduo and Zhong,  Weihe and Lv,  Qiujie and Yu-Chian Chen,  Calvin},
  year = {2022},
  pages = {8693–8703}
}
}

\newpage

 




\vfill

\end{document}